\documentclass[letterpaper]{article} 
\usepackage{aaai25}  
\usepackage{times}  
\usepackage{helvet}  
\usepackage{courier}  
\usepackage[hyphens]{url}  
\usepackage{graphicx} 
\usepackage{natbib}  
\usepackage{caption} 
\usepackage{subcaption}
\usepackage{algorithm}
\usepackage{algorithmic}
\usepackage{amsmath}
\usepackage{amsfonts}
\usepackage{amssymb} 
\usepackage[table,xcdraw]{xcolor}

\usepackage{pifont}

\usepackage{newfloat}
\usepackage{listings}
\DeclareCaptionStyle{ruled}{labelfont=normalfont,labelsep=colon,strut=off} 
\floatstyle{ruled}
\newfloat{listing}{tb}{lst}{}
\floatname{listing}{Listing}

\title{Multimodal Commonsense Knowledge Distillation \\for Visual Question Answering}

\author{
    Shuo Yang\textsuperscript{\rm 1},
    Siwen Luo\textsuperscript{\rm 2},
    Soyeon Caren Han\textsuperscript{\rm 1,2}
}

\affiliations{
    \textsuperscript{\rm 1}University of Melbourne, 
    \textsuperscript{\rm 2}University of Western Australia \\
    syang4418@student.unimelb.edu.au, siwen.luo@uwa.edu.au, caren.han@unimelb.edu.au
}

\begin{document}

\maketitle

\begin{abstract}
Existing Multimodal Large Language Models (MLLMs) and Visual Language Pretrained Models (VLPMs) have shown remarkable performances in general Visual Question Answering (VQA). However, these models struggle with VQA questions that require external commonsense knowledge due to the challenges in generating high-quality prompts and the high computational costs of fine-tuning. In this work, we propose a novel graph-based multimodal commonsense knowledge distillation framework that constructs a unified relational graph over commonsense knowledge, visual objects and questions through a Graph Convolutional Network (GCN) following a teacher-student environment. This proposed framework is flexible with any type of teacher and student models without further fine-tuning, and has achieved competitive performances on the ScienceQA dataset.

\end{abstract}

\section{Introduction}

In recent years, VQA tasks developed to be more challenging by asking questions beyond the image contents and requiring external commonsense knowledge to answer\footnote{Existing VLMs' error cases can be found in Appendix}. Existing works on such commonsense VQA tasks tried different methods to integrate visual, question and commonsense knowledge features \cite{wang2024sco}. For example, \citet{ravi2023vlc} encodes the contextualized commonsense inferences on the question phrases as additional textual features and integrates with object visual features to fine-tune the Vision-and-Language pretrained model (VLPM). VQA-GNN \cite{wang2023vqa} jointly encodes the scene graph of the image and concept graph of the question context as the unified graph for training. T-SciQ \cite{wang2024t} proposes the new chain-of-thought (CoT) prompting strategy to fine-tune the Multimodal large language model (MLLM). However, these works face problems from two aspects: 1) though incorporating CoT in MLLMs has shown remarkable performances on knowledge-based VQA, generating the high-level reasoning CoT is challenging; 2) directly fine-tuning the large VLMs can be computationally expensive. To address these issues, in this work, we propose a multimodal teacher-student knowledge distillation framework that is computationally efficient to jointly learn the features of multi-modalities \cite{cabral20243m, han2020victr, cao2023scenegate}. Specifically, in the teacher model, this framework integrates the object entities from image, question and commonsense knowledge graph together in a unified graph and explicitly learns the relationships among them through the Graph Convolutional Neural Network (GCN)\cite{yao2019graph}, inspired by \citet{han2022understanding} and \citet{long2022gradual}. The learned graph features are passed to the student model, which can be any model structure of a smaller size, for the final answer prediction. Notably, instead of fine-tuning based on one vision-and-language model structure, this framework can be flexibly plugged with any pretrained visual and textual encoder for diverse feature extractions in the teacher model. Moreover, this proposed method provides flexibility that can be adapted to environments with different computational efficacies while maintaining competitive performances compared to large VLPMs and MLLMs. We evaluated our proposed framework with the ScienceQA and achieved competitive results.

\section{Methodology}
\begin{figure}[tbp]
    \centering
    \includegraphics[width=\columnwidth, height = 0.7\linewidth]{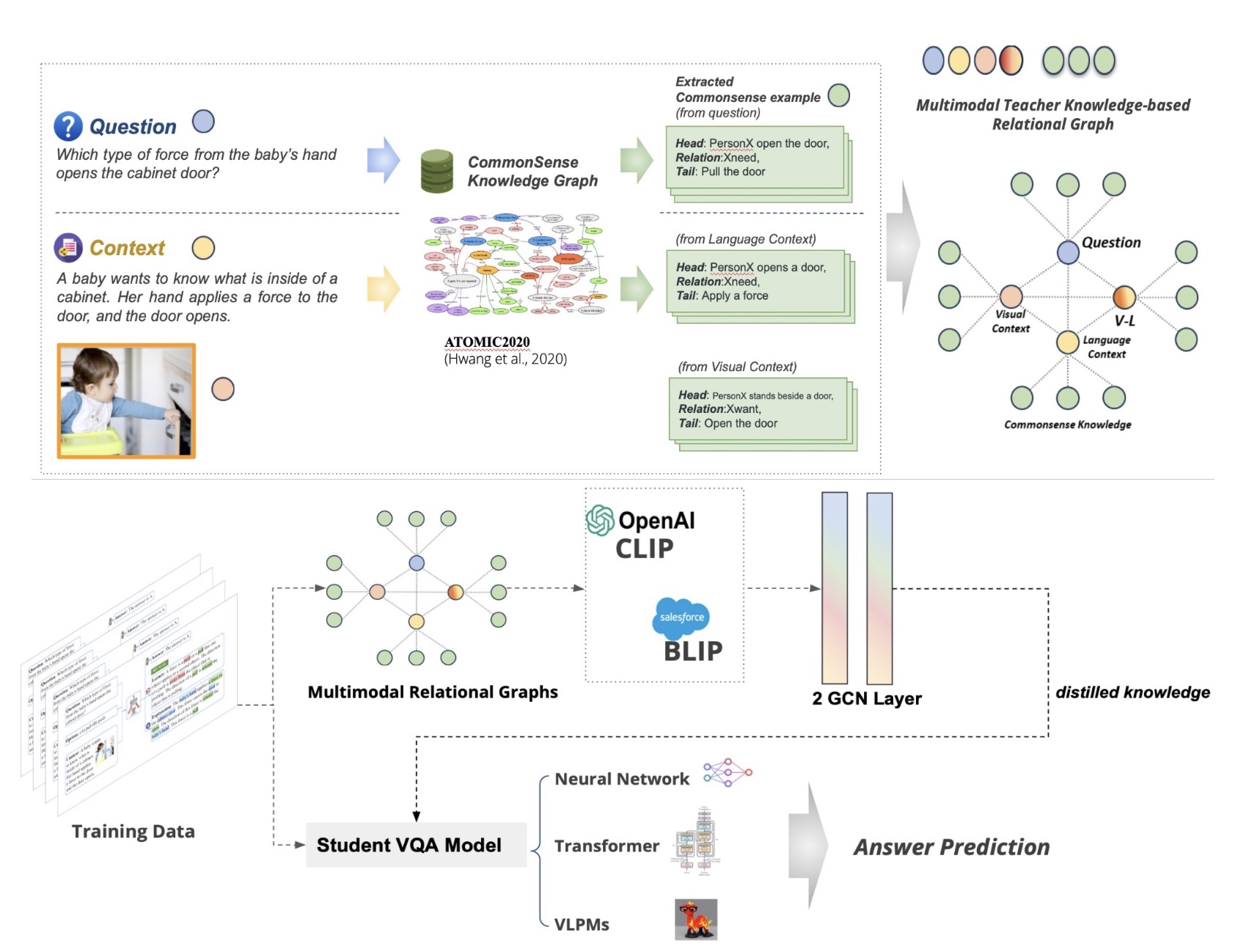}

    \caption{Overall Framework Design}
    \label{fig:Architecture}
\end{figure}

Figure \ref{fig:Architecture} depicts the overall workflow of our proposed graph-based multimodal commonsense knowledge distillation framework. We first represent inputs as graphs to capture the relationships between different modalities enriched by commonsense knowledge. We then employ a GCN to train the teacher graph model. This trained teacher then distils learnt knowledge to the student models of varying size.
 
\textbf{Graph Construction}: To capture the relationships among the multimodal inputs and enrich them with commonsense knowledge understanding, we construct a set of heterogeneous subgraphs \( G = \{G_1, G_2, \ldots, G_M\} \) for a dataset with $M$ samples. Each subgraph $G_{i} = \{V_{i}, E_{i}\}$ represents an individual input sample comprising an image, a question, and contextual information.  The node candidates $V_{i}$ within each subgraph are categorised into two types: content nodes $V_{sub}$ and commonsense nodes $V_{k}$. The content nodes $V_{sub}$ includes four types of node representation for each input modality: a question node for the textual query, a language context node for textual context, a visual context node for image context and a V-L node for combined visual and textual context.

To further inject the model with augmented commonsense knowledge, we integrate commonsense nodes $V_{k}$ into each subgraph. Initially,  each content node $V_{sub}$ is projected into a shared single-modal embedding space using a dual-encoder-based Vision-Language Pretrained Model. We then retrieve relevant commonsense knowledge triplets from the ATOMIC2020 dataset~\cite{hwang2021comet}. Specifically, we compute the cosine similarity between the embedding vector $\mathbf{v}_u$ of each content node $V_u$ and the embeddings $\mathbf{v}_k$ of all triplets in the ATOMIC2020 dataset as illustrated as:
\begin{equation}
\text{sim}(V_u, k) = \frac{\mathbf{v}_u \cdot \mathbf{v}_k}{\| \mathbf{v}_u \| \| \mathbf{v}_k \|}.
\end{equation}
The triplets are pre-embedded into the same shared space using the VLPM. We select the top $K$ triplets with the highest similarity scores for each content node $V_u$ (we set $K = 3$ in our experiments). These selected triplets are considered the most semantically relevant and are added to the subgraph as commonsense nodes $V_k$.

Edges for any pair of nodes  $V_{x}, V_{y} \in V_{sub}$ as well as $V_{u}$ with their retrieved commonsense nodes $V_{k}$ are defined by either Cosine Similarity and Pointwise Mutual Information (PMI). These metrics are chosen to capture semantic relationships and statistical dependencies among the nodes.

\textbf{Graph Learning}: We leverage a standard two-layer Graph Convolutional Network (GCN) to capture the multimodal information and injected commonsense knowledge within the constructed graph. It is illustrated in Equation \ref{equ: GCN}: 
\begin{equation}
     f(V)^{(l+1)} = \sigma^{(l)} \left(\tilde{D}^{-\frac{1}{2}} \tilde{A} \tilde{D}^{-\frac{1}{2}} f(V)^{(l)} W^{(l)}\right)
    \label{equ: GCN}
\end{equation}
where: \(f(V)^{(l)}\in \mathbb{R}^{N \times T^{(l)}} \) represents the node feature at layer \( l \), \( \tilde{A} = A + I_N \in \mathbb{R}^{N \times N} \) is the adjacency matrix of the graph with added self-connections; \( N \) is the number of nodes within each subgraph and \( T^{(l)} \) is the dimension of feature space at layer \( l \). 
We then apply an average pooling \( f_{\text{pooling}}(\cdot) : \mathbb{R}^{N \times T^{(l)}} \to \mathbb{R}^ {1 \times T^{(l)}} \) over each subgraph and feed the pooled embedding over each sub-graph to a multi-layer perception (MLP) \( f_{\text{MLP}}(\cdot) \): $\mathbb{R}^{M \times T^{(L)}} \to \mathbb{R}^{M \times T^{(O)}}$, where $T^{(O)}$ denotes the number of the unique labels. We use the cross-entropy loss to optimize the model. 

\textbf{Multimodal Graph-based Knowledge Distillation}: After training the teacher graph using GCN, we distil soft labels to the student model, where it is optimised by the Kullback-Leibler Divergence (KD) loss as in Equation \ref{equ: KLD}: 
\begin{equation}
\begin{array}{c}
    \mathcal{L}_{KD} = \text{KLDivLoss}\left(\frac{1}{n_T} \sum_{j=1}^{n_T} P_j, P_s\right) \\ 
    P_j = \text{softmax}(T_j(X)) \hspace{0.35cm}
    P_s = \text{softmax}(S(X))
\end{array}
\label{equ: KLD}
\end{equation}
where $T(X) $, $ S(X) $ represents the teacher model and student model. We formulate overall loss by adding up the student cross-entropy loss and KD loss as \(\mathcal{L} = \mathcal{L}_{SCE} +  \mathcal{L}_{KD}\).

\section{Experiments and Results}
We compare the micro F1-score against three types of baseline models of varying size: (1) Small-sized \textbf{MLP};  (2) Medium-sized \textbf{Transformer}; (3) Three Large-sized \textbf{VLPMs} that has been applied in the ScienceQA dataset \cite{lu2022learn}: (a) VisualBERT \cite{li2019visualbert}: integrates RoI-based visual feature and token-based textual feature through BERT-style architecture. (b) ViLT \cite{kim2021vilt}: processes visual and textual tokens using a unified fusion encoder directly. (c) UnifiedQA \cite{khashabi2020unifiedqa}: unifies various QA format throughout a textual-only model. 
We evaluate the proposed framework on the ScienceQA \cite{lu2022learn}. Each group is tested with or without integrating our proposed graph-based knowledge distillation framework. From the overall performance covered in Table \ref{tab: Main Results}, we can see a significant improvement in their average score with 11.21\% and 8.44\% increase separately with our proposed framework for both MLP and Transformer baselines. For large VLPMs, despite their sophistication, we also find a non-trivial increment in their performance. This suggests the robustness and effectiveness of our method.

\begin{table}[t]
\centering
\resizebox{\columnwidth}{!}{%
\footnotesize
\setlength{\tabcolsep}{3pt}
\renewcommand{\arraystretch}{1.0}
\begin{tabular}{lrrr||r}
\hline
\noalign{\hrule height 1.6pt}
\textbf{Model} & \textbf{NAT} & \textbf{SOC} & \textbf{LAN} & \textbf{AVG} \\
\hline
\rowcolor{gray!15} \textit{Small-sized Baseline and Our Result} & & & & \\
MLP                 & 41.21 & 42.33 & 34.11 & 42.71 \\
\textbf{Teacher + MLP (ours)}& \textbf{54.38} & \textbf{49.23} & \textbf{39.28} & \textbf{53.92} \\ 
\hline
\rowcolor{gray!15} \textit{Medium-sized Baseline and Our Result} & & & & \\
Transformer         & 48.44 & 47.15 & 42.72 & 48.35 \\
\textbf{Teacher + Transformer (ours)}& \textbf{57.74} & \textbf{55.35} & \textbf{48.46} & \textbf{56.79} \\ 
\hline
\rowcolor{gray!15} \textit{Large-sized Baselines and Our Result} & & & & \\
ViLT                & 60.48 & 63.89 & 60.27 & 61.14 \\
\textbf{Teacher + ViLT (ours)}    & \textbf{64.12} & \textbf{66.55} & \textbf{63.83} & \textbf{65.41} \\
VisualBERT          & 59.33 & 69.18 & 61.18 & 61.87 \\ 
\textbf{Teacher + VisualBERT (ours)} & \textbf{61.30} &\textbf{ 72.84} & \textbf{64.33} & \textbf{65.69} \\ 
UnifiedQA$_{base}$     & 68.16 & 69.18 & 74.91 & 70.12 \\
\textbf{Teacher + UnifiedQA$_{base}$ (ours)}  & \textbf{71.41} & \textbf{73.22} & \underline{71.58} & \textbf{72.33}\\
\hline
\noalign{\hrule height 1.6pt}
\end{tabular}
}
\caption{Overall Performance on ScienceQA. Question classes: NAT = natural science, SOC = social science, LAN = language science. \cite{lu2022learn}.}
\label{tab: Main Results}
\end{table}

\section{Conclusion}
We proposed a multimodal graph-based commonsense knowledge distillation framework that addresses the limitations of existing VLMs in VQA tasks by integrating object, question, and commonsense knowledge into a unified graph structure and leveraging a GCN for relational learning. Our results on ScienceQA validate the effectiveness of this approach, showing notable performance improvements. 

\newpage

\section{Acknowledgement}
This study was supported by funding from the Google Award for
Inclusion Research Program (G222897). We thank Dr. Josiah Poon (The University of Sydney) and Prof. Eduard Hovy (The University of Melbourne) for fruitful suggestions and discussions.

\bibliography{aaai25}

\begin{thebibliography}{15}
\providecommand{\natexlab}[1]{#1}

\bibitem[{Cabral et~al.(2024)Cabral, Luo, Poon, and Han}]{cabral20243m}
Cabral, R.~C.; Luo, S.; Poon, J.; and Han, S.~C. 2024.
\newblock 3M-Health: Multimodal Multi-Teacher Knowledge Distillation for Mental Health Detection.
\newblock In \emph{Proceedings of the 33rd ACM International Conference on Information and Knowledge Management}, 152--162.

\bibitem[{Cao et~al.(2023)Cao, Luo, Nunez, Wen, Poon, and Han}]{cao2023scenegate}
Cao, F.; Luo, S.; Nunez, F.; Wen, Z.; Poon, J.; and Han, S.~C. 2023.
\newblock Scenegate: Scene-graph based co-attention networks for text visual question answering.
\newblock \emph{Robotics}, 12(4): 114.

\bibitem[{Han et~al.(2020)Han, Long, Luo, Wang, and Poon}]{han2020victr}
Han, C.; Long, S.; Luo, S.; Wang, K.; and Poon, J. 2020.
\newblock VICTR: Visual Information Captured Text Representation for Text-to-Vision Multimodal Tasks.
\newblock In \emph{Proceedings of the 28th International Conference on Computational Linguistics}, 3107--3117.

\bibitem[{Han et~al.(2022)Han, Yuan, Wang, Long, and Poon}]{han2022understanding}
Han, S.~C.; Yuan, Z.; Wang, K.; Long, S.; and Poon, J. 2022.
\newblock Understanding graph convolutional networks for text classification.
\newblock \emph{arXiv preprint arXiv:2203.16060}.

\bibitem[{Hwang et~al.(2021)Hwang, Bhagavatula, Le~Bras, Da, Sakaguchi, Bosselut, and Choi}]{hwang2021comet}
Hwang, J.~D.; Bhagavatula, C.; Le~Bras, R.; Da, J.; Sakaguchi, K.; Bosselut, A.; and Choi, Y. 2021.
\newblock (comet-) atomic 2020: On symbolic and neural commonsense knowledge graphs.
\newblock In \emph{Proceedings of the AAAI conference on artificial intelligence}, volume~35, 6384--6392.

\bibitem[{Khashabi et~al.(2020)Khashabi, Min, Khot, Sabharwal, Tafjord, Clark, and Hajishirzi}]{khashabi2020unifiedqa}
Khashabi, D.; Min, S.; Khot, T.; Sabharwal, A.; Tafjord, O.; Clark, P.; and Hajishirzi, H. 2020.
\newblock UNIFIEDQA: Crossing Format Boundaries with a Single QA System.
\newblock In \emph{Findings of the Association for Computational Linguistics: EMNLP 2020}, 1896--1907.

\bibitem[{Kim, Son, and Kim(2021)}]{kim2021vilt}
Kim, W.; Son, B.; and Kim, I. 2021.
\newblock Vilt: Vision-and-language transformer without convolution or region supervision.
\newblock In \emph{International conference on machine learning}, 5583--5594. PMLR.

\bibitem[{Li et~al.(2019)Li, Yatskar, Yin, Hsieh, and Chang}]{li2019visualbert}
Li, L.~H.; Yatskar, M.; Yin, D.; Hsieh, C.-J.; and Chang, K.-W. 2019.
\newblock Visualbert: A simple and performant baseline for vision and language.
\newblock \emph{arXiv preprint arXiv:1908.03557}.

\bibitem[{Long et~al.(2022)Long, Han, Wan, and Poon}]{long2022gradual}
Long, S.; Han, S.~C.; Wan, X.; and Poon, J. 2022.
\newblock Gradual: Graph-based dual-modal representation for image-text matching.
\newblock In \emph{Proceedings of the IEEE/CVF winter conference on applications of computer vision}, 3459--3468.

\bibitem[{Lu et~al.(2022)Lu, Mishra, Xia, Qiu, Chang, Zhu, Tafjord, Clark, and Kalyan}]{lu2022learn}
Lu, P.; Mishra, S.; Xia, T.; Qiu, L.; Chang, K.-W.; Zhu, S.-C.; Tafjord, O.; Clark, P.; and Kalyan, A. 2022.
\newblock Learn to explain: Multimodal reasoning via thought chains for science question answering.
\newblock \emph{Advances in Neural Information Processing Systems}, 35: 2507--2521.

\bibitem[{Ravi et~al.(2023)Ravi, Chinchure, Sigal, Liao, and Shwartz}]{ravi2023vlc}
Ravi, S.; Chinchure, A.; Sigal, L.; Liao, R.; and Shwartz, V. 2023.
\newblock Vlc-bert: Visual question answering with contextualized commonsense knowledge.
\newblock In \emph{Proceedings of the IEEE/CVF winter conference on applications of computer vision}, 1155--1165.

\bibitem[{Wang, Han, and Poon(2024)}]{wang2024sco}
Wang, E.; Han, C.; and Poon, J. 2024.
\newblock SCO-VIST: Social Interaction Commonsense Knowledge-based Visual Storytelling.
\newblock In \emph{Proceedings of the 18th Conference of the European Chapter of the Association for Computational Linguistics (Volume 1: Long Papers)}, 1602--1616.

\bibitem[{Wang et~al.(2024)Wang, Hu, He, Xu, Liu, Liu, and Shen}]{wang2024t}
Wang, L.; Hu, Y.; He, J.; Xu, X.; Liu, N.; Liu, H.; and Shen, H.~T. 2024.
\newblock T-sciq: Teaching multimodal chain-of-thought reasoning via large language model signals for science question answering.
\newblock In \emph{Proceedings of the AAAI Conference on Artificial Intelligence}, volume~38, 19162--19170.

\bibitem[{Wang et~al.(2023)Wang, Yasunaga, Ren, Wada, and Leskovec}]{wang2023vqa}
Wang, Y.; Yasunaga, M.; Ren, H.; Wada, S.; and Leskovec, J. 2023.
\newblock Vqa-gnn: Reasoning with multimodal knowledge via graph neural networks for visual question answering.
\newblock In \emph{Proceedings of the IEEE/CVF International Conference on Computer Vision}, 21582--21592.

\bibitem[{Yao, Mao, and Luo(2019)}]{yao2019graph}
Yao, L.; Mao, C.; and Luo, Y. 2019.
\newblock Graph convolutional networks for text classification.
\newblock In \emph{Proceedings of the AAAI conference on artificial intelligence}, volume~33, 7370--7377.

\end{thebibliography}

\end{document}